# Predicting the Long-Term Outcomes of Biologics in Psoriasis Patients Using Machine Learning

S. Emam[1], A.X. Du[2], P. Surmanowicz[2], S.F. Thomsen[3], R. Greiner[4,5], R. Gniadecki[2*]

[1]Information Services and Technology, University of Alberta, Edmonton, Alberta, Canada;
[2]Division of Dermatology, Faculty of Medicine and Dentistry, University of Alberta, Edmonton, Alberta, Canada; [3]Department of Dermatology, Bispebjerg Hospital, University of Copenhagen, Denmark; [4] Alberta Machine Intelligence Institute (Amii); Edmonton, Alberta, Canada, [5]Computing Science, Faculty of Science, University of Alberta, Edmonton, Canda

*corresponding author: *Robert Gniadecki, MD, Division of Dermatology, University of Alberta, 8-112 Clinical Science Building, 11350-83 Ave, Edmonton, AB, T6G 2G3, Canada*

*Tel: (780) 407-1555, Fax: (780) 407-2996, email: [r.gniadecki@ualberta.ca](mailto:r.gniadecki@ualberta.ca)*

## Summary

*Background*. Real-world data show that approximately 50% of psoriasis patients treated with a biologic agent will discontinue the drug because of loss of efficacy. History of previous therapy with another biologic, female sex and obesity were identified as predictors of drug discontinuations, but their individual predictive value is low.

*Objectives*. To determine whether machine learning algorithms can produce models that can accurately predict outcomes of biologic therapy in psoriasis on individual patient level.

*Results.* All tested machine learning algorithms could accurately predict the risk of drug discontinuation and its cause (e.g. lack of efficacy vs adverse event). The learned generalized linear model achieved diagnostic accuracy of 82%, requiring under 2 seconds per patient using the psoriasis patients dataset. Input optimization analysis established a profile of a patient who has best chances of long-term treatment success: biologic-naive patient under 49 years, early-onset plaque psoriasis without psoriatic arthritis, weight < 100 kg, and moderate-to-severe psoriasis activity (DLQI ≥ 16; PASI ≥ 10).  Moreover, a different generalized linear model is used to predict the length of treatment for each patient with mean absolute error (MAE) of 4.5 months. However Pearson Correlation Coefficient indicates 0.935 linear dependencies between the actual treatment lengths and predicted ones.

*Conclusions*. Machine learning algorithms predict the risk of drug discontinuation and treatment duration with accuracy exceeding 80%, based on a small set of predictive variables. This approach can be used as a decision making tool, communicating expected outcomes to the patient, and development of evidence-based guidelines.

## Introduction

The unprecedented success and wide implementation of biologics in the therapy of psoriasis has over the last decade changed the landscape of the medical need in this disease. In the pre-biologic era, skin clearance was difficult to achieve and required a combination of multiple skin-directed therapies and systemic agents, with a significant risk of cumulative toxicities.

Today, biologics allow us to achieve PASI 75 responses in up to 90% of the patients, and real-world evidence from registries confirms that a comparable proportion of psoriasis patients achieve excellent control of skin disease.[1,2] Despite the vast amount of data on the efficacy of biologics, therapeutic decision-making (ie, deciding which treatment to administer for each individual patient) is still based on a trial-and-error approach. The initial choice of therapy is not always optimal, as reflected by data documenting that over 50% of patients need dose optimisation during the therapy and 20%-50% of patients experience relapse of the disease and require a switch to another medication.[1,3–5]

There is clearly a need for a personalized medicine approach that would allow for a more accurate prediction of the appropriate choice of the drug and dose at the initial assessment, and communicate chances of long-term success to an individual patient. Several studies addressed this issue using multivariate logistic regression to identify possible predictors of outcome. It seems that obesity, female sex, failure of another biologic therapy in the past, presence of psoriasis on palms and soles and socioeconomic status significantly correlate with worse long-term outcomes.[1,4,6] Unfortunately, the positive predictive values for these factors are very low and they are rarely incorporated into clinical decision making.

We have considered the possibility that the long-term efficacy of biologics is to a certain extent a predictable phenomenon and depends on subtle, complex patterns of interactions between numerous variables. Machine learning techniques are often able to detect such complex patterns and are being increasingly used with success in predicting future trajectories of patients' health in diverse areas such as cancer,[7] the risk of readmission after hospital discharge,[8] diabetic complications,[9,10] cardiovascular mortality,[11] and many others.[12] Although machine learning is frequently employed for drug discovery,[13] its use for predicting long-term outcomes in patients in real-world setting has not been widely investigated.

Here we examined whether machine learning may aid in predicting long-term responses to biologics in psoriasis. We have reanalysed individual patient data from the Danish registry cohort, Dermbio, using drug discontinuation as a surrogate measure of treatment failure.[1,4,5]

## Methods

### *Dataset*

We have used individual patient data from the Danish registry, Dermbio, of 681 psoriasis patients who received biologic treatment in one of five academic centers in Denmark between 2003 and 2013. The structure and the data available in Dermbio have been previously described in a detail.[1,4,5,14] All patient data were anonymized. Therapy outcome measures are the length of the treatment series and the cause of drug discontinuation at the last observation categorized as being caused by: lack of efficacy, adverse event(s), other (patient's decision, loss to follow-up, not otherwise specified). This dataset has been described in detail in our previous publication.[4]

Patient characteristics and available variables used for machine learning algorithms are listed in **Table 1**.

**Table 1.** Patient characteristics and variables used in the machine learning algorithms

| Category | | Data completeness (%) |
|---|---|---|
| **Demographics** | | |
| *Age (years)* | 42.8 (9, 83)[1] | 100 |
| *Sex* | Male: 375<br>Female: 306 | 100 |
| *Height (cm)* | 174.1 (110, 198)[1] | 74.74 |
| *Weight (kg)* | 85.6 (30, 180)[1] | 57.27 |
| *Number of comorbidities* | 0: 395<br>1: 136<br>2: 40<br>≥3: 22 | 87.08 |
| **Disease-specific** | | |
| *Age at diagnosis (years)* | 25.84 (9, 70)[1] | 80.32 |
| *PsA diagnosis* | No: 454<br>Yes: 227 | 100 |
| *Previous MTX* | No: 543<br>Yes: 138 | 100 |
| *Concurrent MTX* | No: 319<br>Yes: 49 | 54.04 |
| *Previous biologic use* | 0: 464<br>1 or more: 217 | 100 |
| *Baseline DLQI* | 13.56 (0, 32)[1] | 38.18 |
| *Baseline PASI* | 10.5 (0, 39.4) | 8.25 |
| **Therapy-specific** | | |
| *Biologic initiated* | Adalimumab: 253<br>Etanercept: 196<br>Infliximab: 117<br>Ustekinumab: 115 | 100 |
| *Treatment series in same patient* | No: 248<br>Yes: 433 | 100 |

[1] mean (range)

### *Data preprocessing and feature engineering*

In order to apply different machine learning algorithms on psoriasis patients, we performed the following preprocessing steps: (1) Correct data types/formats were applied to all features and meaningful naming conventions were applied to feature names; (2) Features with large number of missing values were removed from the training dataset (e.g. PASI after 13 and 52 weeks, DLQI after 13 and 52 weeks); (3) Principal Components Analysis (PCA) was applied to detect and remove non-predictive features; (4) Predictors which were still missing values were filled with Null or appropriate value indicating unavailability of the data.

*Machine Learning*

Understanding the reasons for biologic discontinuation, different supervised machine learning techniques and algorithms were used to extract a stochastic model from the patients' dataset and used to predict outcomes of biologic therapy in psoriasis patients whose treatment outcome is unknown. In this study, we considered seven different modelling techniques, based upon their principally different prediction approaches: (1) Generalized Linear Model (GLM), (2) Logistic Regression,(3) Deep Learning, (4) Decision Tree (DT), (5) Random Forest, (6) Gradient Boosted Trees, and (7) Support Vector Machine (glm2 package in R, GLM and DT packages in Python, RapidMiner simulator).

We applied different machine learning algorithms on the same dataset to evaluating the performance of different approaches and select the learner that performed the best in terms of accuracy, interpretability and runtime in a multinomial classification problem. However, as GLM and DT outperforms other learners in terms of accuracy, we describe them in more detail in the following sections. In order to assess the effectiveness of the models, we applied the 5-fold cross validation technique as part of the performance analysis procedure. This technique randomly partitions a set of examples into 5 (non-overlapping) sets. Subsequently, a model is inferred using each set over 5 iterations and the remaining set is used to evaluate the model based upon the accuracy measure (defined in the next section). Since in each iteration a different set is used for the evaluation, the final accuracy score is equal to the mean of the 5 accuracy scores. This assists in evaluating the performance of our machine learning approach on some unseen data and identifying over-fitting or under-fitting related issues.

*Performance analysis*

In order to statistically evaluate the performance of learning approaches used in this study, we produced confusion tables as a result of classification procedure. A confusion table (matrix) reports the number of TP: True Positives; FN: False Negatives, TN: True Negatives and FP: False Positives, which can be used to calculate diagnostic accuracy with the following formula:

$$Accuracy = \frac{TP+TN}{TP+FP+FN+TN}$$

Receiver operating characteristic (ROC) curves were constructed using R libraries multiROC and ggplot2 (downloaded from https://cran.r-project.org/web/packages/) to visualise diagnostic utility of the machine learning algorithms. Area under the curve (AUC) was calculated to determine which model provided the best prediction by measuring how true positive rate (recall) and false positive rate trade off. It is worth noting that the ROC-AUCs were calculated by keeping a class and stacking the rest of the groups together, thus converting the multi-class classification into binary classification.

The Bland-Altman analysis[15] was used to compare the result of prediction on the length of treatment series versus the actual treatment length.

# Results

## *Machine learning accurately predicts the risk of discontinuation*

Drug persistence is often used as a surrogate measure of treatment success in a real-world setting because most of biologic drug discontinuation events are due to loss of the therapeutic efficacy or adverse events.[1,3,4] We have applied seven different machine learning algorithms to test their accuracy in predicting the risk of drug discontinuation. As shown in **Table 2**, all algorithms predicted the risk of discontinuation with high accuracy ranging from 0.73 (support vector machine learning) to 0.82 (generalized linear model, GLM). Thus, we were able to predict the treatment outcomes with less than 18.46% classification error, just utilizing the basic health information routinely available to every clinician.

**Table 2.** Performance of different machine learning algorithms in predicting the risk of discontinuation of the biologic in patients with psoriasis.

| Model | Accuracy | Standard Deviation | Runtime (s) |
|---|---|---|---|
| Generalized Linear Model | 0.815 | 0.028 | 1.584 |
| Logistic Regression | 0.759 | 0.047 | 3.144 |
| Deep Learning | 0.754 | 0.043 | 2.608 |
| Decision Tree | 0.795 | 0.044 | 1.413 |
| Random Forest | 0.805 | 0.039 | 17.541 |
| Gradient Boosted Trees | 0.810 | 0.047 | 53.394 |
| Support Vector Machine | 0.733 | 0.059 | 12.55 |

## *Generalized Linear Model (GLM)*

Generalized linear models (GLMs) are an extension of traditional linear models that considers the response variables with arbitrary error distribution models. This algorithm fits generalized linear models to the data by maximizing the log-likelihood and using iteratively reweighted least squares method. GLM first was formulated by McCullagh and Nelder[16] to gather different statistical models (e.g. linear regression, logistic regression and Poisson regression) under one umbrella. In GLM the computation process for model fitting is parallel, extremely fast, and scales extremely well for models with a limited number of predictors with non-zero coefficients.

To better visualize the usefulness of GLM for predicting the risk of discontinuation, we plotted the sensitivity and specificity achieved by this algorithm as a ROC curve (**Fig 1**). The GLM algorithm was not only able to accurately predict the overall risk of discontinuation, but also robustly predicted the reason of discontinuation, lack of efficacy versus adverse events. The confusion matrix describing the performance of the classification model on the set of labeled data is available in **supplementary Figure S1**.

*Input optimization reveals patient profile associated with the best long-term response*

Input optimization (or prescriptive analytics) in machine learning describes a process through which the algorithm finds a set of characteristics that best fits the desired future outcome. In other words, it describes the optimal scenario to correctly determine the classification results for unseen inputs. Therefore, using input optimization suitable actions can be suggested to benefit from prediction and show the implication of different decision options.

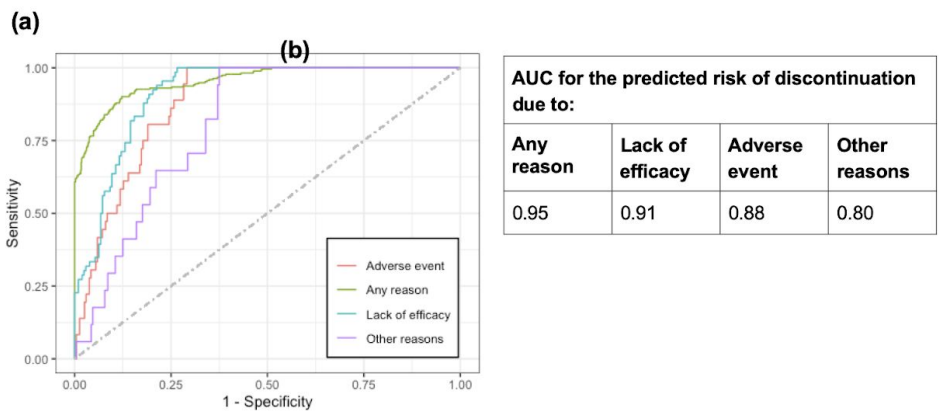

**Fig. 1. Multiple Class ROC and AUC Analysis for the Generalized Linear Model (GLM). (a)** The curves show the performance of the model for predicting drug discontinuation due to any reason, adverse event, lack of efficacy, or other reasons (patient decision, loss to follow up, not specified). **(b)** AUC values for different causes of drug discontinuation.

Here, we were interested in knowing the characteristics under which the patient would continue the treatment with less than 10% chance of withdrawal. The results indicated that a patient who has at least 90% chance of continuing the treatments fulfill the following criteria: (1) ≥ 23 year-old at the time of diagnosis (2) ≤ 49 year-old at the time of treatment; (3) receiving ustekinumab rather than TNF inhibitor, (4) not diagnosed with psoriasis arthritis (5) baseline DLQI ≥ 16; (6) baseline PASI ≥ 9.4; (7) no previous history of biologic failure, (8) weight ≤ 98.9 kg.

*Decision tree analysis*

The decision tree analysis yielded a slightly lower accuracy in predicting the risk of discontinuation compared to GLM (**Table 2** and **supplementary Fig S2**). However, the advantage of the decision tree is the ability to generate interpretable results. A decision tree is a flowchart-like collection of nodes intended to predict the value of target variable (in this study, treatment discontinuation) using input variables (see **Table 1**, Methods). Each node represents a splitting rule for one specific variable. Depending on the prediction goals, this rule can either separate values belonging to different classes or separate them in order to reduce the error in an optimal way for the selected parameter criterion. The building of new nodes is repeated until the stopping criteria are met. The decision tree algorithm repeatedly selects the branches of the tree to obtain subsets of data that contain a larger proportion of the same value of the target variable comparing the level above it. Thus, the algorithm determines the appropriate tree structure by recursive partitioning, or repeated splitting on the values of attributes until an endpoint is reached. An overview of the inferred tree-like model for psoriasis patients is shown in **supplementary Figure S3**. It shows that the length of the treatment has a meaningful effect on the treatment discontinuation rate. For instance, the model revealed that the discontinuation that occurs in patients who receive the biologic for less than a year is most likely to be due to inefficacy whereas adverse events play an important role in those who are treated for more than a year. The algorithm has also highlighted that patients treated with infliximab compared to other biologics are more likely to withdraw the treatment due to inefficacy within the first two years of starting the treatment regardless of age and gender. These results, although confirmatory, underscore the ability of the model to extract meaningful patterns from our dataset.

*Predicting the treatment length*

From the practical point of view it is useful to be able not only to predict the risk of treatment termination, but also the length of the therapy until the predicted drug discontinuation. We have compared the predicted treatment length obtained by the GLM model to the actual recorded values using Bland Altman analysis[15] (**Fig 2**). The Bland Altman approach quantifies the agreement between the two measurements in terms of limits of agreement for their mean difference. As shown in **Fig 2A** the GLA model slightly underestimated the length of treatment for patients who stayed on therapy for less than 40 months and overestimated the treatment duration for treatment duration >40 months. However, the overall correlation between the predicted and the actual treatment duration was excellent (**Fig 2B**) with the mean absolute error (MAE) of only 4.5 months.

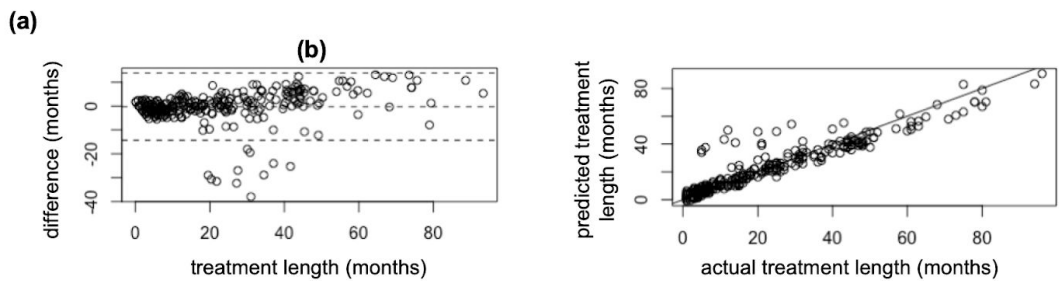

**Fig. 2. Bland Altman analysis of the agreement between GLM and actual length of therapy. (a)** The mean between the predicted and actual length of treatment ([months], x-axis) is plotted against the difference between these values (y-axis). The dotted line shows 95% confidence intervals. **(b)** Correlation plot between the actual and predicted treatment length.

**Discussion**

Machine learning is emerging as a powerful tool to improve diagnostic accuracy and prediction of treatment outcomes in clinical medicine.[12] Here, for the first time we provide evidence that machine learning is able to accurately predict the treatment success in psoriasis patients treated with biologics. In particular, the GLM model accurately predicted the overall risk of drug discontinuation with an accuracy exceeding 80% and was also very efficient in forecasting the cause of discontinuation as the loss of efficacy or adverse events. The GLM also accurately determined the expected length of treatment with the mean absolute error (MAE) of 4.5 months.

The very high accuracy of GLM was surprising, taking into account a relatively small dataset with a limited set of predictive variables. The dataset from Dermbio only offered 14 different variables, all of those easily available in a real world clinical setting (**Table 1**). The simplicity of the data structure and high computational efficiency of our algorithms (runtime ranging between 1.4 to 53 sec on a commercially available laptop) would allow to implement our machine learning algorithms in any clinical setting with minimal additional cost. The GLM can easily be expanded to include more relevant data for even higher predictive accuracy. It is conceivable, than feeding the model with additional information, such as the dose of the biologic, the actual characteristics of comorbidities (rather than their number), concomitant medication and the clinical characteristics of psoriasis would significantly improve its usefulness. Even with this very limited dataset we were able to confirm the importance of obesity and previous failure of a biologic as negative predictors of treatment success.[1,4,5,17] Interestingly, the previously seen negative impact of female sex[6,17] was not detected if the baseline PASI and DLQI data were included in the training set. Also, the impact of PsA was negative, in contrast to what has previously been suggested from real-world data[3,6,18] and in a recent metaanalysis.[17] Our model led to a profile of the ideal patient who would most likely benefit from biologic therapy long-term: a patient with early-onset psoriasis (before the age of 23 years), the age under 49 years when receiving a biologic for the first time, no diagnosis of PsA, and who has a moderate-to-severe psoriasis activity (DLQI ≥ 16; PASI ≥ 10) and weight <100 kg. The overall risk of drug discontinuation in such a patient is ≤10% which is clearly superior to the mean of the general population which is approximately 20% after the first year of treatment and 50% after 3 years.[4,6,17]

An important advantage of the machine learning models, such as GLM is that they yield results in a format which is immediately understandable for the patient and the clinician. Classic approach with logistic regression or Cox proportional hazards regression analyses produce odds ratios and hazard ratios, respectively, as measures of treatment effect. These are often erroneously interpreted as being synonymous with relative risks.[19] However, the decrease in hazard ratio or odds ratio does not correspond to the same percentage of clinical improvement, which makes those measures very difficult to use for the patient communication purposes. Although odds ratios can in some instances be converted to relative risks,[20] patients prefer to receive information as absolute values, such as the expected number of years to the event or the absolute risk of positive or negative outcome.[21,22] GLM generates such values, the risk of drug discontinuation and the projected time on successful therapy, which can conveniently be presented as a personalized risk simulator. This approach resembles cardiovascular risk calculators widely used in general medicine and cardiology to demonstrate the benefits of therapeutic intervention.[23–25]

It is conceivable that machine learning would not only be useful as a tool of precision medicine but would also aid in the guideline development. There are numerous psoriasis registries collecting data on the outcomes of biologic therapies that together comprise over 60,000 patients.[14] In spite of the efforts such as PsoNet, the analysis of the totality of registry data has not been accomplished, mostly due to differences in the structure of different databases, differences in the registration of the outcome and lack of statistical tool to analyse the very diverse datasets. The machine learning approach would be a viable approach to analyse the long-term outcomes in psoriasis and provide quantitative data informing health care providers and policy makers.

In addition to the already mentioned small sample size, the limitation of this study is the source of data from a single country. As extensively discussed elsewhere, the duration of treatment of biologics is not only dependent on the objective responses but may be heavily influenced by non-medical factors such as access to the medication, reimbursement, or treatment guidelines. Thus, the accuracy of our algorithm may vary between different centers and each therapy center should establish their own algorithm which reflects local prescription practices. Another limitation of this study is its retrospective nature. This is a usual practice in machine learning modeling that a dataset is divided into a smaller training set which is used to generate the model and a test set used to validate the model. However, the ideal proof of usefulness would be through a prospective study in which the choice of treatment performed by the machine learning algorithm is compared with the outcomes achieved by the decision of the physician.

## Acknowledgements

A.X.D. would like to thank Alberta Innovates and the Canadian Association of Psoriasis Patients for providing her with a summer research stipends, which supported her in conducting this work. The study was funded from the start-up grant from the Department of Medicine, University of Alberta to R.Gn.

## Funding Sources

This study was funded by a start-up grant from the Department of Medicine, University of Alberta to R.Gn. A.X.D. was supported by summer studentships from Alberta Innovates and the Canadian Association of Psoriasis Patients.

## Conflict of Interest Disclosure

S.E., A.X.D., P.S., and R.Gr. report no conflicts. S.F.T. has been a paid speaker for AbbVie, Eli Lilly, Novartis, Sanofi, Pierre Fabre, GSK and LEO Pharma, and has served on Advisory Boards with AbbVie, Eli Lilly, Janssen, Novartis, Roche, Sanofi, UCB, and LEO Pharma. He has served as an investigator for AbbVie, AstraZeneca, Boehringer, UCB, CSL and Novartis and received research grants from AbbVie, Novartis, Sanofi and UCB. R.Gn. reports carrying out clinical trials for AbbVie and Janssen and has received honoraria as a consultant and/or speaker from AbbVie, Bausch Health, Eli Lilly, Janssen, Mallincrodt, Novartis, and Sanofi. The authors do not have equity in pharmaceutical companies.

|  | Reason for drug discontinuation | TRUE | | | | | |
|---|---|---|---|---|---|---|---|
|  |  | Adverse event | Patient's decision | Lack of efficacy | Loss to follow up | Other | Continue |
| **PREDICTED** | Adverse event | 6 | 3 | 4 | 0 | 1 | 0 |
|  | Patient's decision | 0 | 0 | 0 | 0 | 0 | 0 |
|  | Lack of efficacy | 17 | 4 | 45 | 2 | 3 | 2 |
|  | Loss to follow up | 0 | 0 | 0 | 0 | 0 | 0 |
|  | Other | 0 | 0 | 0 | 0 | 0 | 0 |
|  | Continue | 0 | 0 | 0 | 0 | 0 | 108 |

Performance vector: accuracy: 81.54% +/- 2.81% (micro average: 81.54%), classification error: 18.46% +/- 2.81% (micro average: 18.46%)

**Figure S1. Confusion matrix for predicting the risk of drug discontinuation for GLM.**

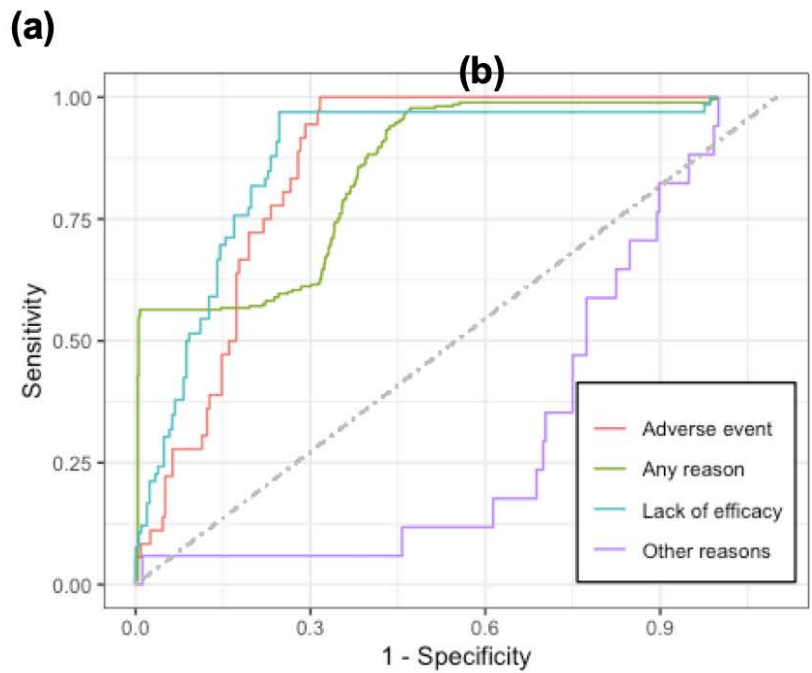

| AUC for the predicted risk of discontinuation due to: | | | |
|---|---|---|---|
| Any reason | Lack of efficacy | Adverse event | Other reasons |
| 0.83 | 0.87 | 0.84 | 0.26 |

**Figure S2. Multiple Class ROC and AUC Analysis for the Decision Tree Model. (a)** The curves show the performance of the model for predicting drug discontinuation due to any reason, adverse event, lack of efficacy, or other reasons (patient decision, loss to follow up, not specified). **(b)** AUC values for different causes of drug discontinuation.

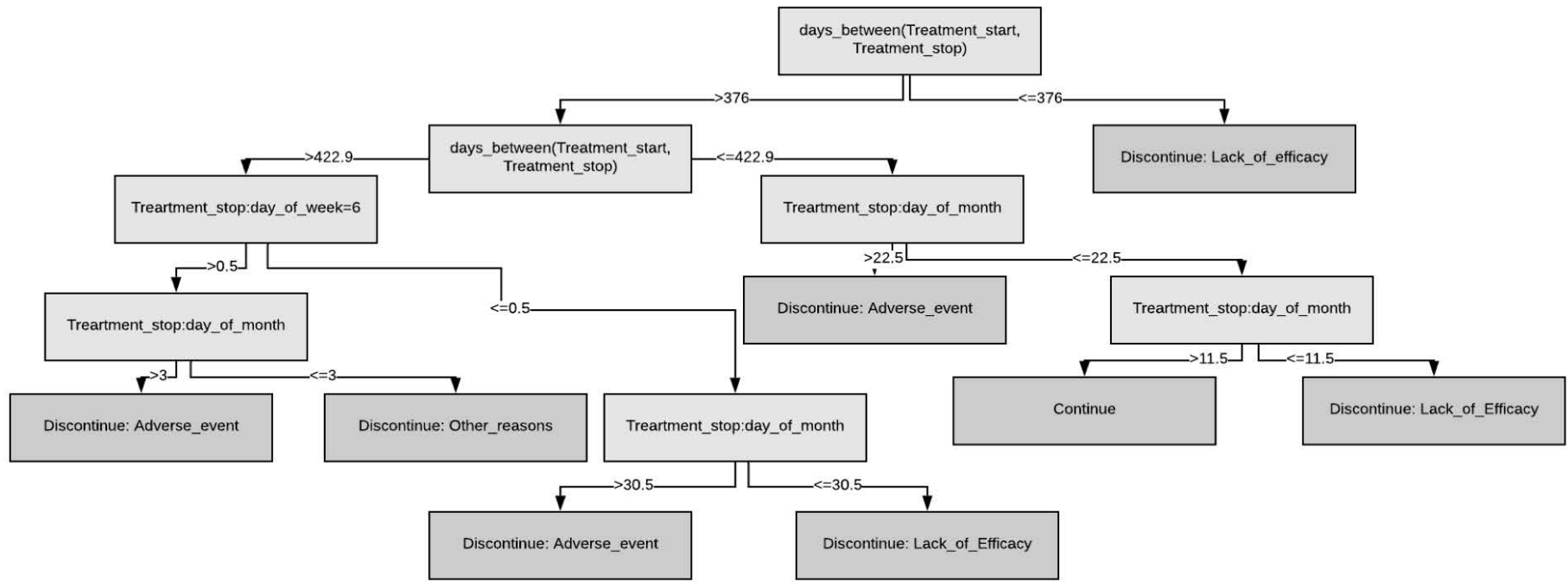

**Figure S3. The decision tree for the analysis of drug discontinuation**.